\documentclass{article} % For LaTeX2e

    \newcommand{\redacted}[1]{[link removed for anonymity]}
    \newcommand{\hide}[1]{}
\usepackage{iclr2026_conference}

\usepackage{iclr2026_conference,times}

\usepackage{amsmath,amsfonts,bm}

\def\eqref#1{equation~\ref{#1}}
\def\1{\bm{1}}

\DeclareMathAlphabet{\mathsfit}{\encodingdefault}{\sfdefault}{m}{sl}
\SetMathAlphabet{\mathsfit}{bold}{\encodingdefault}{\sfdefault}{bx}{n}

\usepackage{hyperref}
\usepackage{url}

\usepackage[utf8]{inputenc} % UTF-8 support
\usepackage[T1]{fontenc}    % 8-bit T1 fonts
\usepackage{amsmath}
\usepackage{cleveref}  % \Cref for \ref
\usepackage{listings}  % syntax highlighting
\usepackage{subcaption}
\usepackage{multicol}
\usepackage{booktabs}
\usepackage{enumitem}

\definecolor{keywordcolor}{rgb}{0.7, 0.1, 0.1}   % red
\definecolor{tacticcolor}{rgb}{0.0, 0.1, 0.6}    % blue
\definecolor{commentcolor}{rgb}{0.4, 0.4, 0.4}   % grey
\definecolor{symbolcolor}{rgb}{0.0, 0.1, 0.6}    % blue
\definecolor{sortcolor}{rgb}{0.1, 0.5, 0.1}      % green
\definecolor{attributecolor}{rgb}{0.7, 0.1, 0.1} % red
\def\sl{\emph{Sledgehammer}\xspace}
\def\lh{\textsc{LeanHammer}\xspace}
\def\lp{\textsc{LeanPremise}\xspace}

\usepackage{pifont}  % provides \ding

\newcommand{\cmark}{\ding{51}}%
\definecolor{xmark}{RGB}{175,175,175}
\newcommand{\xmark}{\textcolor{xmark}{\ding{55}}}%

\usepackage{graphicx}  % resizebox for wide tables
\usepackage{adjustbox}  % adjustbox for wide tables
\usepackage{siunitx}  % aligning, rounding decimals

\usepackage{xspace} % for automatic space after a macro command
\newcommand{\miniCTX}{\texttt{miniCTX}\xspace}

\newcommand{\modelsizeone}{\texttt{small}\xspace}
\newcommand{\modelsizetwo}{\texttt{medium}\xspace}
\newcommand{\modelsizethree}{\texttt{large}\xspace}
\newcommand{\settingauto}{\texttt{auto}\xspace}
\newcommand{\settingaesop}{\texttt{aesop}\xspace}
\newcommand{\settingaesopauto}{\texttt{aesop+auto}\xspace}
\newcommand{\settingfull}{\texttt{full}\xspace}
\newcommand{\settingcumul}{\texttt{cumul}\xspace}

\usepackage{amsfonts}
\title{Premise Selection for a Lean Hammer}

\author{
Thomas Zhu\textsuperscript{1,\textasteriskcentered},
Joshua Clune\textsuperscript{1,\textasteriskcentered}, \\
\;\textbf{Jeremy Avigad\textsuperscript{1,\textdagger},
Albert Q.\ Jiang\textsuperscript{2,\textdagger},
Sean Welleck\textsuperscript{1,\textdagger}} \\
\textsuperscript{1}Carnegie Mellon University, \textsuperscript{2}Mistral AI \\
\textsuperscript{\textasteriskcentered}Equal contribution, \textsuperscript{\textdagger}Equal advising \\
\texttt{\{thomaszh,jclune,avigad,swelleck\}@andrew.cmu.edu, qj213@cam.ac.uk}
}

\iclrfinalcopy % Uncomment for camera-ready version, but NOT for submission.
\begin{document}

\maketitle

\begin{abstract}
    Neural methods are transforming automated reasoning for proof assistants, yet integrating these advances into practical verification workflows remains challenging. A \emph{hammer} is a tool that integrates premise selection, translation to external automatic theorem provers, and proof reconstruction into one overarching tool to automate tedious reasoning steps. 
    We present \textsc{LeanPremise}, a novel neural premise selection system, and we combine it with existing translation and proof reconstruction components to create \lh, the first end-to-end domain general hammer for the Lean proof assistant. Unlike existing Lean premise selectors, \lp is specifically trained for use with a hammer in dependent type theory. It also dynamically adapts to user-specific contexts, enabling it to effectively recommend premises from libraries outside \lp's training data as well as lemmas defined by the user locally. With comprehensive evaluations, we show that \lp enables \lh to solve 21\% more goals than existing premise selectors and generalizes well to diverse domains. Our work helps bridge the gap between neural retrieval and symbolic reasoning, making formal verification more accessible to researchers and practitioners.\footnote{\lp is available at \url{https://github.com/hanwenzhu/premise-selection} and \lh is available at \url{https://github.com/JOSHCLUNE/LeanHammer}.}
\end{abstract}

\section{Introduction}

Interactive proof assistants have long been used to verify the correctness of hardware, software, network protocols, cryptographic protocols, and other computational artifacts. Buoyed by successes like the Liquid Tensor Experiment \citep{LeanCommunityLTE} and the formalization of the Sphere Eversion Theorem \citep{VanDoornEtAl2023SphereEversion}, mathematicians are increasingly using the technology to verify mathematical theorems \citep{Tao2023FPRBluePrint} and build substantial mathematical libraries \citep{Mathlib}.

When working with a proof assistant, a user describes a proof in an idealized proof language, which is a programming language that provides sufficient detail for the computer to construct a precise formal derivation in the proof assistant's underlying axiomatic system. One of the challenges to formalization is the requirement to spell out what seem like straightforward inferences in painful detail. This problem is exacerbated by the fact that at the most basic level of interaction, users are required to name the required premises (i.e., definitions and lemmas) explicitly to justify an inference step, from a library of hundreds of thousands of previously derived facts.

A \textit{hammer} \citep{MengEtAl2006FirstSledgehammer,PaulsonBlanchette2012threeYears,blanchette2016hammering} is a tool designed to ease the pain of formalization by filling in small inferences automatically. Typically, a hammer has three components: given a goal to prove, one first selects a moderate number of premises from the library, project files, current file, and hypotheses that, one hopes, are sufficient to prove the goal. This is known as \emph{premise selection}. Then one translates the premises and the goal into the language of powerful external automated theorem provers like Vampire \citep{KovacsVoronkov2013Vampire}, E \citep{SchultzEtAl2019Eprover}, and Zipperposition \citep{VukmirovicEtAl2022Zipperposition}, or SMT solvers like Z3 \citep{DeMouraBjorner2008Z3} and cvc5 \citep{BarbosaEtAl2022Cvc5}. Finally, if the external prover succeeds in proving the goal, it reports back the specific premises used, from which a formal proof in the proof assistant is reconstructed.
% sometimes by redoing the search with a less powerful but proof-producing method.

In this paper, we present \lp, a new premise selection tool for the Lean proof assistant \citep{DeMouraUllrich2021Lean}. We combine it with the DTT-to-HOL~(dependent type theory to higher-order logic) translation tool, Lean-auto \citep{qian2025lean-auto}, and internal proof-producing tactics, Duper \citep{CluneEtAl2024Duper} and Aesop \citep{limperg2023aesop}, resulting in \lh, the first end-to-end domain general hammer for Lean. Through comprehensive evaluations, we show that \lh can hit nails.

Our work, which extends methods of premise selection used by Magnushammer \citep{mikula2024magnushammer} and LeanDojo \citep{yang2023leandojo}, is therefore an auspicious combination of neural premise selection methods with symbolic proof search. For the first time, we specifically design contrastive learning methods for the first end-to-end domain general hammer in Lean. We explain the design choices to make \lp performant for \lh, including new \textit{hammer-aware data extraction} techniques. An important feature of \lp is that it dynamically augments the library of facts with locally defined facts from the user's project, which is essential in practice.

Our core contributions are as follows:
\begin{itemize}[leftmargin=*]
\item We develop \lp, a premise selection tool for a hammer in dependent type theory.
\item We combine \lp with Aesop, Lean-auto, and Duper to make \lh, the first domain general hammer in Lean.
\item We provide an accessible user-facing tactic interface that can dynamically process new premises in the environment.
\item We conduct comprehensive evaluations of \lh's performance on Mathlib and its ability to generalize to \miniCTX-v2 \citep{hu2025minictx}. Through these evaluations, we show that \lh solves 21\% more goals with \lp than with existing premise selectors and that \lp enables \lh to effectively use libraries and premises it hasn't seen before.
\end{itemize}

Note that premise selection can be used in other ways, for example, for calling various types of internal automation directly, for presenting suggestions to a user engaged in manual proof, or for use in a neural or neurosymbolic search. Although our focus here has been on a hammer, we expect that many of the methods we develop carry over to other settings.

\section{Related work}
\label{sec:related:work}

\subsection{Hammers in interactive proof assistants}

As explained in the introduction, hammers support interactive proving by completing small inferences, called \emph{goals}.
The first and still most successful hammer in use today is Isabelle's Sledgehammer, developed initially by \cite{MengEtAl2006FirstSledgehammer} and further developed by \cite{PaulsonBlanchette2012threeYears, BlanchetteEtAl2013SledgehammerSMT}, and many others. Since then hammers have been developed for HOL \citep{KaliszykUrban2015HOLyHammer}, Mizar \citep{KaliszykUrban2015MizarHammer}, Rocq \citep{CzajkaKaliszyk2018CoqHammer}, and Metamath \citep{CarneiroEtAl2023MetamathHammer}, among others. Of these, only Rocq is based on dependent type theory. Despite Lean's popularity, no hammer has been developed for Lean.

\subsection{Neural theorem proving}
\label{subsection:neural}

Numerous neural-network-based tools have been developed to prove theorems. A straightforward approach of using neural models is to let them generate steps in proofs, notable examples of which include GPT-\textit{f}~\citep{polu2023gptflean}, HTPS~\citep{lample2022htps}, ReProver~\citep{yang2023leandojo}, DeepSeek-Prover~\citep{xin2024dsprover,xin2024dsprover15} for Lean, LISA~\citep{jiang2021lisa} and Thor~\citep{jiang2022thor} for Isabelle, and PALM~\citep{lu2024palm}, Cobblestone~\citep{kasibatla2024cobblestone}, and Graph2Tac~\citep{graph2tac} for Coq/Rocq. Another line of work uses neural models to generate entire proofs or proof sketches~\citep{JiangEtAl2023DraftSketchProve, zhao2023decomposing, wang2024lego, first2023baldur, lin2025goedel, lin2025goedel-v2, wang2025kimina, chen2025seed}. These proof search approaches are complementary to a hammer, which serves as a tactic that may be used by neural models.

Hammers are embedded in a number of neural theorem proving frameworks such as Thor and Draft, Sketch, and Prove~\citep{jiang2022thor, zhao2023decomposing, JiangEtAl2023DraftSketchProve, wang2024lego} to fill small gaps in the proofs. It is worth noticing that all these works use the Isabelle proof assistant~\citep{NipkowEtAl2002Isabelle} where the communication infrastructure~\citep{jiang2021lisa} between neural models and the proof assistant is relatively mature and hammering is easy to set up.
Our work makes calling a hammer in Lean possible.
% \todo{(Albert) Is this an overstatement? (Jeremy) Sounds good to me.}

% Hammers can also be used in conjunction with neural theorem methods. For example, the Draft-Sketch-Prove architecture \citep{JiangEtAl2023DraftSketchProve} proved theorems in the Isabelle proof assistant \citep{NipkowEtAl2002Isabelle} by turning informal proofs into formal proof sketches and then calling a hammer to complete the remaining goals. In the Thor framework \citep{JiangEtAl2022Thor}, an LLM was trained to generate calls to a hammer while generating a proof incrementally, among other possible proof steps.

Despite a large number of research works, practical tools that a working mathematician has access to without complex setup or prohibitive costs remain scarce. Recent state-of-the-art methods use reinforcement learning on e.g.\ 7B LLMs with thousands of passes for a single theorem and use infrastructures not callable from Lean~\citep{wu2024internlm2, lin2025goedel, lin2025goedel-v2, dong2025beyond, chen2025seed}, so it is prohibitive for Lean users to train, test, or use them. Our work brings forward a tool that is packaged as a tactic and can be called straightforwardly from any IDE for Lean with low computational cost and latency, hence enabling better automation for the masses.

\subsection{Premise selection}

Formalizing mathematics in proof assistants requires users to select relevant premises from libraries of hundreds of thousands of facts. To help facilitate this task, premise selection has been developed for a variety of proof assistants, using both neural and symbolic techniques.
MePo \citep{meng2009lightweight_mepo} is a symbolic premise selector which has been widely used in Isabelle's Sledgehammer. Other premise selectors which target hammers but use traditional machine learning techniques include MaSh \citep{kuhlwein2013mash}, $k$-NN based premise selection for HOL4 \citep{Gauthier2015-rp}, CoqHammer's premise selection \citep{CzajkaKaliszyk2018CoqHammer}, and random forest based premise selection for Lean \citep{piotrowski2023machine}. 
\lp differs from these by using modern LM-based retrieval methods.

(L)LM-based premise selection trained by contrastive learning has also been explored for a variety of use cases. Lean State Search \citep{Tao2025-cc} recommends relevant premises directly to Lean users. Magnushammer \citep{mikula2024magnushammer} generates premises to supply directly to proof reconstruction tactics. ReProver and Lean Copilot \citep{yang2023leandojo, song2025leancopilotlargelanguage} retrieve premises to augment neural next-tactic generation. Unlike these, \lp is specifically designed with hammer integration in mind, which requires specific data extraction and loss formulation, and the resulting selector to be fast and domain general.

\section{Methods}

\subsection{\lh pipeline} \label{sec:hammer-pipeline}
Hammers broadly consist of three primary components: premise selection, translation to external automatic theorem provers, and proof reconstruction. In traditional hammer pipelines, such as Isabelle's Sledgehammer, these components are composed in a linear fashion, with the premises from premise selection informing the translation to automatic theorem provers and the output from automatic theorem provers informing proof reconstruction. In other works,
such as Magnushammer \citep{mikula2024magnushammer} and Lean Copilot \citep{song2025leancopilotlargelanguage}, premises from premise selection
are provided directly to proof reconstruction tactics or language models, without translating and sending them to external automatic
theorem provers. 
\lh introduces a new, unified hammer pipeline that uses premise selection in both of these ways.
% Informed by the experimental results discussed in \Cref{sec:results}, \lh uses premise selection in both of these ways.

\begin{figure}[ht]
    \centering
    \includegraphics[width=0.9\linewidth]{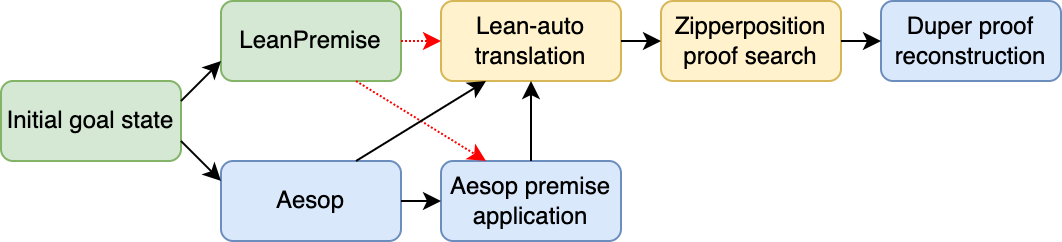}
    \caption{Overview of the \lh pipeline. Phases that can neither fail nor produce a terminal proof are
    green, phases that can fail but cannot produce a terminal proof are yellow, and phases that can produce a
    terminal proof are blue. Black solid arrows indicate control flow, while red dashed
    arrows indicate the transfer of information between phases.}
    \label{fig:hammer-outline}
\end{figure}

\Cref{fig:hammer-outline} gives an overview of the \lh pipeline. In addition to \lp itself, \lh is built upon Aesop, Lean-auto, and Duper. Aesop is a highly extensible proof search tool that can be  augmented with new proof search rules and procedures. Lean-auto
is a translation tool that does not search for proofs itself, but instead translates dependently typed
Lean goals into higher-order logic problems which can be solved by external automatic theorem provers such as
Zipperposition. Finally, Duper is a less powerful but proof-producing proof search tool which implements many of the techniques found in
automatic theorem provers, and is therefore well suited to rediscovering and verifying proofs found by
external automatic theorem provers.

In broad strokes, Aesop is called
first and prioritizes finding a proof using its own built-in rules. If a short proof using only built-in rules
is not found quickly, it explores direct premise applications using premises recommended by \lp,\footnote{Premise applications are
rules added to Aesop of the form \texttt{(add unsafe 20\% <premise>)} where \texttt{<premise>} is a
premise selected by the premise selector.}
and it queries Lean-auto to see if subgoals can be closed using premises from the selector.\footnote{Lean-auto
is added to Aesop as a rule of the form \texttt{(add unsafe 10\% (by auto [*, <premises>]))} where
\texttt{<premises>} is a list of premises selected by the premise selector.} When
Lean-auto is given a subgoal, it translates that subgoal (along with the premises provided by the selector)
to higher-order logic and interfaces with Zipperposition to find a proof. If Zipperposition succeeds, then
Duper is provided just the set of premises used by Zipperposition to solve the translated problem, and Duper attempts
to reconstruct a proof from these premises. For an illustrative example of \lh's pipeline in action, see \Cref{sec:leanhammer-example}.

\subsection{Data extraction}
\label{sec:data-extraction}

To support \lp, we develop a data extraction pipeline designed to gather not just information useful for next-tactic generation or human examination, but all of the information that may be helpful for a hammer tasked with discovering an end-to-end proof. This pipeline is used dynamically to extract premises that \lp can retrieve at runtime, including premises or definitions defined by the user locally, and it is used statically to extract (state, premise) pairs for training. As we describe our data extraction pipeline, we note the measures taken to collect data that go beyond what appears explicitly in the source code for the formal proofs.

\subsubsection{Signature extraction}
\label{sec:signature-extraction}

A key aspect of premise selectors is how premises are presented to the model. Previous work \citep{yang2023leandojo} extracts raw strings from the source code, which ignores many details in the full signature (see \Cref{sec:data-example} for an example). We adopt a new \emph{normalized serialization} as follows. For each theorem and definition in each module,
we extract the documentation description in the source code (its docstring), if it exists, as well as its kind (theorem or definition), name, arguments, and overall type. 
Together, these can be composed into a signature of the form \texttt{docstring? kind name arguments* : type}.
When converting these signatures into strings, we disable notation pretty printing (e.g.~we
print $\mathbb{N}$ as \verb|Nat|), and we print every constant with its fully qualified name (e.g.~we
print \verb|I| as \verb|Complex.I|). This standardizes premise representation, so that it depends only on the type of the premise and does not depend on open namespaces, custom notations, and surface-level syntax, which may change at run time. For an illustrative example, see \Cref{sec:data-example}.

% For example, de Moivre's formula is extracted as:

% \begin{lstlisting}[language=lean, basicstyle=\ttfamily\fontsize{8pt}{8pt}\selectfont]
%     /-- **De Moivre's formula** -/
%     theorem Complex.cos_add_sin_mul_I_pow (n : Nat) (z : Complex) : Eq (HPow.hPow (HAdd.hAdd (Complex.cos z) (HMul.hMul (Complex.sin z) Complex.I)) n) (HAdd.hAdd (Complex.cos (HMul.hMul (↑n) z)) (HMul.hMul (Complex.sin (HMul.hMul (↑n) z)) Complex.I))
% \end{lstlisting}

The signatures extracted in this manner are used to form the set of premises $\mathcal{P}$ that \lp is allowed to retrieve from. This signature extraction pipeline is also used to dynamically extract new premises at runtime (\Cref{sec:api-integration}). To prevent \lp from 
constantly recommending theorems that are technically relevant to the goal but never useful for our hammer's 
automation, we filter out a blacklist of 479 basic logic theorems such as \verb|and_true| from $\mathcal{P}$.
We also filter out Lean language-related (e.g.\ metaprogramming) definitions not useful for proofs.

\subsubsection{State and premise extraction}
\label{sec:state-premise-extraction}

The next key question is which (state, premise) pairs are extracted from human-written proofs to train the model. Previous premise selectors \citep{yang2023leandojo} that focus on tactic generation only extract from tactic-style proofs, and only extract explicit premises appearing in the raw source code of only the next tactic. Our \textit{hammer-aware data extraction} improves upon this in several ways. First, we extract from both term-style and tactic-style proofs, significantly increasing training samples especially for short proofs that \lh is intended to automate. Second, for multi-tactic proofs, the model is trained to select premises to close the goal (all tactics) rather than to modify the goal (first tactic), because hammers are designed to finish proofs. Third, we extract both implicit and explicit premises from the proof, including ones implicitly called by automation such as \verb|simp|. Finally, we format states with the same normalized serialization as for premises.

Specifically, for each theorem in each module, we collect data on the premises used to prove it. Additionally,
for each theorem proven via tactic-style proofs, we collect data on all intermediate
goal states induced by the tactic sequence. For an illustrative example, see \Cref{sec:data-example}. Ultimately, all data we extract contains:
\begin{itemize}[leftmargin=*]
    \item A proof state obtained either from the beginning of a theorem or from an intermediate step of a
    tactic-style proof.
    \item The name and signature of the theorem from which the state was extracted.
    \item The set of premises used to prove the theorem\footnote{We also experimented with 
    pairing states with just the set of premises used to close said states, as opposed to all premises used to 
    prove the overall theorem, but our preliminary experiments showed that this yielded worse results than 
    including all premises.}.
\end{itemize}

% Sean: This seems a key/important part of getting the premise selector to work well and is novel, but I feel that most readers will get lost in the details or skip these two paragraphs. I am not sure what to do yet, but just noting it
% Unlike previous work in tactic or whole-proof generation, a hammer typically needs all premises that are useful for closing the current goal, so we extract the full set of implicit and explicit premises that contribute to a proof from all types of proofs.
When theorems are proven via term-style proofs, meaning the theorem's proof term is 
explicitly written in the source code, the set of premises we extract is the set of theorems that appear in 
the proof term.
When theorems are proven via tactic-style proofs, meaning automation is invoked to tell Lean how to build
a proof term, the set of premises we extract contains both the theorems that appear in the proof term constructed
by the tactic sequence (so that all implicit premises are collected), as well as any theorems and definitions that are explicitly used in \verb|rw| or
\verb|simp| calls.

The benefit of collecting explicit theorems and definitions from \verb|rw| and \verb|simp| calls relates to
Lean's dependent type theory. In Lean, terms can be definitionally equal without being syntactically equal,
and because of this, tactic-style proofs can invoke definitional equalities that do not appear in
final proof terms. We therefore collect these definitional equality premises.
We experimentally verify that our hammer-aware data extraction benefits \lh in \Cref{sec:ablation}.

\subsection{Premise selection}
\label{sec:premise-selection}

\lp uses the standard method of retrieval using textual encoders. In order to retrieve $k$ premises for a state $s$, we first determine the set $\mathcal{P}_s$ of accessible premises at position $s$, comprising lemmas and definitions that are imported from other modules or declared earlier in the file. We use an encoder-only transformer model $E$ to embed both the state $s$ and every premise $p\in \mathcal{P}_s$, and the resulting set of premises retrieved is
\begin{equation}
    \texttt{select\_premises}(s, k, \mathcal{P}_s) = \text{top-$k$}_{p \in \mathcal{P}_s}\texttt{sim}(E(s), E(p))
\end{equation}
where $\texttt{sim}(u,v)=u^\top v/\|u\|_2\|v\|_2$ is cosine similarity. In \Cref{sec:api-integration} we describe the mechanism for caching embedding and quick retrieval of the premises.

We do not train a separate reranking model as in \citet{mikula2024magnushammer}, because we did not find it to increase performance in early experiments, especially since a hammer favors recall much more than precision, and we determined the optimal $k$ to be at least $16$, at which point reranking does not offer much improvement. It is also costly to deploy in practice.

\subsubsection{Model training} \label{sec:model-training}
We use a modified version of the InfoNCE loss~\citep{oord2018representation} to train the encoder model.
On a high level, each batch consists of (state, premise) pairs, and a contrastive loss is used to let the model learn to select the correct premise out of all premises in this batch. One problem is that there are many premises in the library that do not appear in any proof. This is mitigated by also sampling negative premises in each batch \citep{mikula2024magnushammer, yang2023leandojo}. Another problem is that there are many premises that are shared across many proofs, so not all premises in the batch are negative. We use the following masked contrastive loss to address these problems.

Specifically, for each training step, we sample a batch of $B$ (state, premise) pairs, each consisting of a state $s_i$ and a premise $p_i^+\in \mathcal{P}_{s_i}^+$ where $\mathcal{P}_{s_i}^+$ is the set of ground-truth premises for $s_i$ extracted as in \Cref{sec:data-extraction}.
For each such pair $(s_i,p_i^+)$, we sample $B^-$ negative premises $\{p_{ij}^-\}_{j=1}^{B^-} \subseteq \mathcal{P}_{s_i} \setminus \mathcal{P}_{s_i}^+$, giving $B$ states and $B(1+B^-)$ premises in total in each batch. Of these premises, we determine the set $\mathcal{N}_i = \{p_i^+\}_i \cup \{p_{ij}^-\}_{ij} \setminus \mathcal{P}_{s_i}^+$ of negative premises for state $s_i$, and mask out the positive ones in the loss to avoid mislabeling. The loss is:
\begin{align}
    \mathcal{L}(E) = \frac{1}{B}\sum_{i=1}^{B}\frac{\exp(\texttt{sim}(E(s_i),E(p_i^+))/\tau)}{\exp(\texttt{sim}(E(s_i),E(p_i^+))/\tau) + \sum_{p_i^-\in \mathcal{N}_i}\exp(\texttt{sim}(E(s_i),E(p_i^-))/\tau)}
    \label{eq:loss}
\end{align}
where $\tau$ is a scalar  temperature hyper-parameter (set to 0.05 in our experiments).
% We found a large $B=256$ improves model performance.

\subsubsection{API integration} % (what is a good name?)
\label{sec:api-integration}

\begin{table}[tb]
\small
\centering
\adjustbox{max width=\linewidth}{
\begin{tabular}{lccccc}
\toprule
\textbf{Premise selector} & \textbf{LM-based} & \textbf{Callable in Lean} & \textbf{New premises} \\
\midrule
ReProver \citep{yang2023leandojo}        & \cmark & \xmark & \xmark \\
Lean Copilot \citep{song2025leancopilotlargelanguage}      & \cmark & \cmark & \xmark \\
Random forest \citep{piotrowski2023machine}            & \xmark & \cmark & \xmark \\
MePo \citep{meng2009lightweight_mepo}    & \xmark & \cmark & \cmark \\
\midrule
\textbf{\lp} & \cmark & \cmark & \cmark \\
\bottomrule
\end{tabular}
}

% {\footnotesize *An unoptimized prototype of MePo was recently adapted into Lean by Kim Morrison at \url{https://github.com/leanprover/lean4/tree/mepo}.}
\caption{Usability comparison of existing premise selection tools. Note that this is orthogonal to the quantitative performance comparisons (\Cref{tab:performance}).} \label{tab:qualitative-comparison}
\end{table}

In order to make \lp and \lh more accessible for Lean users as well as downstream methods, we design our pipeline to maximize usability---it is directly callable in Lean, able to take in new premises, and efficient to run.
Our pipeline for premise selection is as follows: when a user invokes premise selection, the client side (Lean) collects all currently defined premises $\mathcal{P}_s$ defined in the environment and the current proof state $s$ and sends them to a server that hosts the embedding model. The server embeds both the proof state and the list of premises, and then runs FAISS \citep{douze2024faiss} on the premises to compute $\texttt{select\_premises}(s,k,\mathcal{P}_s)$, and returns this list of $k$ premises back to the client.
Since the typical size of $\mathcal{P}_s$ is on the scale of $\sim$70k, the server also caches the embeddings of premises at fixed versions of Mathlib, and only recomputes embeddings of signatures of new premises uploaded by the user (e.g.\ when working outside Mathlib or when the user has new premises in the context); the client side also caches the signatures of these new premises computed as in \Cref{sec:signature-extraction}.

\lh is built as a tactic that can be directly called in Lean. It calls \lp as a subprocedure and the retrieved premises are then input to the \lh pipeline (\Cref{sec:hammer-pipeline}).
In Mathlib, premise selection usually takes about 1 second amortized on a CPU server (and well under 1 second for a single-GPU server). The full \lh pipeline on average takes well under 10 seconds (see \Cref{sec:results}).

To the best of our knowledge, \lp is the first premise selector using language models that can be directly invoked in Lean and can incorporate new user-defined premises. It is also efficient to run and requires no system setup for the user, because the main computation is only a few string embeddings, and done centrally in a server by default. This makes the premise selector itself a desirable user-facing tactic for the Lean community. Similarly, the full \lh can be called straightforwardly in Lean as a tactic. This bridges a gap that many previous LM-based retrievers and provers leave. See \Cref{tab:qualitative-comparison} for a comparison.

\subsection{Variations and extensions} \label{sec:variations}
Here, we discuss variations on \lh's design, implemented as settings that can be controlled by the user. Note that the pipeline described in \Cref{sec:hammer-pipeline} has premises input both to Aesop as premise application rules and to Lean-auto for
translation to the external prover. We consider variants that disable either one:

\begin{enumerate}[leftmargin=*]
    \item \settingaesop: This setting only inputs \lp's premises to Aesop as premise applications, omitting calls to Lean-auto or the external prover.
    \item \settingauto: This setting inputs \lp's premises directly to the external prover through Lean-auto without Aesop normalization or premise application.
    \item \settingaesopauto: This setting keeps both Aesop and Lean-auto, but does not use premise applications as Aesop rules.
    \item \settingfull: This default setting is the full pipeline described in \Cref{sec:hammer-pipeline}.
\end{enumerate}
These variants are appealing because they offer cheaper computational cost while still preserving much of \lh's ability. Experiments offer insight to the ability of each part of the pipeline (see \Cref{sec:results}). We observe that the first three variants may prove theorems that \settingfull does not, so we also consider \settingcumul, which tries all four variants.

We note that other common domain-general automation tactics that take premises as inputs, such as \verb|simp_all|, may be roughly considered a subcase of \settingaesop (which we verify in preliminary experiments), so we do not consider them. We also tried using a second-stage model to predict \verb|simp_all| ``hints''---whether a premise should be supplied to \verb|simp_all| for preprocessing, and whether it should be applied in reverse direction, but the performance did not increase.
We remark that additional automated reasoning tactics in the future may be easily added to our pipeline as a rule of Aesop, similarly to how Lean-auto is added.

% comment on the originally proposed tests:
% hammer_nosimp actually performs better than hammer (with simp_all []), because (1) simp_all fails on some theorems (max heartbeat) and (2) hammerCore itself improved a lot since we last considered hammer (with simp_all []). Furthermore, hammer (with simp_all []) is basically a subcase of aesop+hammerCore because of aesop's normalization. So we don't consider the default setting of hammer (with simp_all [])

\section{Experiments}

\subsection{Experimental setup}
We extract theorem proofs from Mathlib, and premises from Mathlib, Batteries, and Lean core. In total, we extract 469,965 states from 206,005 theorem proofs, and extract 265,348 (filtered) premises. For each state, there are on average 12.45 relevant premises, giving 5,817,740 (state, premise) pairs in the training set. We randomly hold out 500/500 theorems as valid/test sets, respectively.

We train the model from three base models that were pre-trained for general natural language embedding tasks \citep{reimers-2019-sentence-bert}. These are \modelsizeone\footnote{https://huggingface.co/sentence-transformers/all-MiniLM-L6-v2} with 6 layers and hidden size 384 trained from MiniLM-L6, \modelsizetwo\footnote{https://huggingface.co/sentence-transformers/all-MiniLM-L12-v2} with 12 and 384 from MiniLM-L12, and \modelsizethree\footnote{https://huggingface.co/sentence-transformers/all-distilroberta-v1} with 6 and 768 from DistilRoBERTa-base, respectively.
We train our models with learning rate $2e{-4}$, $B=256$, and $B^-=3$, found by a hyperparameter sweep. Training the \modelsizethree model requires 6.5 A6000-days.

We test \lh on proving theorems in (1) our hold-out sets extracted from Mathlib, and (2) the non-Mathlib splits of \miniCTX-v2-test \citep{hu2025minictx}.
We impose a 10-second time constraint for each call to Zipperposition, and for each theorem a 300-second wall-clock time-out and Lean's default heartbeat limit of 200,000. We tuned the value of $k$ on Mathlib-valid (see \Cref{sec:finding-k}), and \settingfull uses the highest performing combination, which is $k_1=16$ premises supplied to Lean-auto (with Aesop priority $10\%$) and $k_2=32$ premises for premise application rules (with Aesop priority $20\%$).
Similarly, we use $k=16$ for \settingauto and \settingaesopauto, and $k=32$ for \settingaesop.

For all experiments and data extraction tasks, we use Lean version v4.16.0. We run a maximum of 16 parallel tests on 16 CPUs with 512GB total memory, so 1 CPU and 32GB are allocated per test theorem. In practice, the actual memory used rarely exceeds 4GB. Each CPU is AMD EPYC 9354 (3.8GHz, 32 cores, 64 threads) or similar.

\subsection{Results}
\label{sec:results}

For each theorem, we record the average percentage of ground-truth premises retrieved in the top-$k$ premises (\emph{recall@$k$}), and the percentage of theorems proven (\emph{proof rate}), shown in \Cref{tab:performance,tab:minictx-performance}. We favor recall over metrics like precision, because a hammer can tolerate irrelevant premises much more than missing important ones.

\begin{table}[!tbp]
\small
% control whether display valid or test performance
\newcommand{\validortest}[2]{#2}

\centering
\adjustbox{max width=\linewidth}{
% rounding and aligning numbers
\sisetup{
  round-mode=places, round-precision=1,  % 1 after .
  table-number-alignment=center,
  detect-weight=true, % detect-inline-weight=math % for bold
}
\begin{tabular}{
    lc
    S[table-format=2.1, table-align-text-post=false] % for * to align correctly
    S[table-format=2.1, table-align-text-post=false] % for * to align correctly
    S[table-format=2.1]
    S[table-format=2.1]
    S[table-format=2.1]
    S[table-format=2.1]
    S[table-format=2.1]
}
\toprule
\textbf{Premise selector} & \textbf{Model size} & \multicolumn{2}{c}{\textbf{Recall} (\%)} & \multicolumn{5}{c}{\textbf{Proof rate} (\%)} \\
\cmidrule(lr){3-4} \cmidrule(l){5-9}
&& {@16} & {@32} & {\settingaesop\unskip} & {\settingauto\unskip} & {\settingaesopauto\unskip} & {\settingfull\unskip} & {\settingcumul\unskip} \\
\midrule
% \cmidrule(r){1-2}
% \cmidrule(lr){3-4}
% \cmidrule(l){5-9}
None
& ---
& \validortest{0}{0}
& \validortest{0}{0}
& \validortest{}{16.9}
& \validortest{}{9.4}
& \validortest{}{16.9}
& \validortest{}{16.9}
& \validortest{}{16.9}
\\
% $k$-NN* % \citep{piotrowski2023machine}
% &
% & \validortest{}{27.567155067155063}
% & \validortest{}{29.04979117705307}
% & \validortest{}{25.6158}
% & \validortest{}{15.2709}
% & \validortest{}{26.1084}
% & \validortest{}{26.1084}
% & \validortest{}{27.0936}
% \\
Random forest* % \citep{piotrowski2023machine}
& --- % RF: https://huggingface.co/hanwenzhu/wip-lean-embeddings/blob/main/train-and-predict-rf.extracted.%2Ball.%2Bn.%2Bb.23685.forest
& \validortest{}{22.1}
& \validortest{}{22.3}
& \validortest{}{19.1}
& \validortest{}{11.9}
& \validortest{}{19.1}
& \validortest{}{19.1}
& \validortest{}{19.1}
\\
MePo % \citep{}
& ---
& \validortest{}{38.4}
& \validortest{}{42.1}
& \validortest{}{23.3}
& \validortest{}{14.5}
& \validortest{}{21.5}
& \validortest{}{26.3}
& \validortest{}{27.5}
\\
ReProver % \citet{yang2023leandojo}
& 218M % NB: 218M is bare encoder for retriever, I think full encoder+decoder is 300M?
& \validortest{36.8}{35.1}\textsuperscript{\textdagger}
& \validortest{41.6}{38.7}\textsuperscript{\textdagger}
& \validortest{12.8}{11.4}
& \validortest{13.8}{12.9}
& \validortest{}{20.5}
& \validortest{14.2}{12.0}
& \validortest{}{22.3}
\\
\midrule
\textbf{\lp} (\modelsizeone)
& 23M
& \validortest{62.2}{59.2}
& \validortest{71.6}{67.8}
& \validortest{28.0}{23.9}
& \validortest{\bfseries 22.8}{19.1}
& \validortest{}{25.9}
& \validortest{34.4}{27.9}
& \validortest{}{31.9}
\\
\textbf{\lp} (\modelsizetwo)
& 33M
& \validortest{61.1}{58.6}
& \validortest{71.9}{68.1}
& \validortest{\bfseries 29.8}{23.1}
& \validortest{22.6}{20.1}
& \validortest{}{26.1}
& \validortest{\bfseries 34.6}{28.5}
& \validortest{\bfseries ?}{30.7}
\\
\textbf{\lp} (\modelsizethree)
& 82M
& \validortest{\bfseries 66.1}{\bfseries 63.5}
& \validortest{\bfseries 75.9}{\bfseries 72.7}
& \validortest{28.6}{\bfseries 24.1}
& \validortest{22.2}{\bfseries 21.3}
& \validortest{}{\bfseries 28.5}
& \validortest{34.4}{\bfseries 30.1}
& \validortest{}{\bfseries 33.3}
\\
\midrule
\multicolumn{2}{l}{\textbf{\lp} (\modelsizethree) $\cup$ MePo}
&
&
& 28.9
& 23.9
& 30.3
& 35.9
& 37.6
\\
% multicolumn because the second column isn't needed and we can take its space
\multicolumn{2}{l}{\textbf{\lp} (cumulative)}  % cumulative over model size (does NOT include +naive data row; ablations should be separate table)
& 
& 
& \validortest{31.0}{27.5}
& \validortest{28.8}{25.5}
& \validortest{}{31.1}
& \validortest{40.0}{34.5} % NB: for valid, xsmall U small is 38.0
& \validortest{}{37.3}
\\
\multicolumn{2}{l}{\textbf{\lp} (cumulative) $\cup$ MePo}
&
&
& 30.3
& 27.1
& 32.3
& 38.2
& 39.6
\\
\midrule
\multicolumn{2}{l}{Ground truth}
& 
& 
& \validortest{30.8}{27.7}
& \validortest{32.0}{33.1}
& \validortest{}{37.8}
& \validortest{41.2}{41.0}
& \validortest{}{43.0}
\\
\bottomrule
\end{tabular}
%% archived results: for leandojo/xs/s/m/cumul/gt/gt-naive, the cumul over aesop/hammer/aesop+hammer without aesop+hammer2 is
%% \validortest{20.8}{18.0723} \validortest{36.8}{30.7229} \validortest{\bfseries 37.4}{29.7189} \validortest{37.2}{\bfseries 31.9277} \validortest{42.2}{36.5462} \validortest{43.6}{42.9719} \validortest{42.4000}{}
}
{\scriptsize *Performance upper bound, excluding errors. \; \textsuperscript{\textdagger}Our definition is slightly different from \citet{yang2023leandojo}. \; See \Cref{sec:baseline-settings}.}
\caption{Performance of \lh with different premise selectors on Mathlib-\validortest{valid}{test}.} \label{tab:performance}

\end{table}

\begin{table}[!tbp]
\small
\centering
\adjustbox{max width=\linewidth}{
\sisetup{
  round-mode=places, round-precision=1,  % 1 after .
  table-number-alignment=center,
  detect-weight=true, % detect-inline-weight=math % for bold
}
\begin{tabular}{
    l
    S[table-format=2.1]
    S[table-format=2.1]
    S[table-format=2.1]
    S[table-format=2.1]
    S[table-format=2.1]
    S[table-format=2.1]
    S[table-format=2.1]
}
\toprule
\textbf{Premise selector} &
\multicolumn{7}{c}{\textbf{Proof rate using \settingfull} (\%)} \\
\cmidrule(l){2-8}
& {Carleson} & {ConNF} & {FLT} & {Foundation} & {HepLean} & {Seymour} & {Average} \\
\midrule
None
& 0.0 & 10.0 & 27.3 & 38.0 & 8.0 & 6.0 & 14.9
\\
\textbf{\lp} (\modelsizethree)
& 0.0 & 16.0 & 36.4 & 38.0 & 10.0 & 24.0 & 20.7
\\
Ground truth
& 7.1 & 16.0 & 39.4 & 40.0 & 20.0 & 34.0 & 26.1
\\
\bottomrule
\end{tabular}
}
\caption{Out-of-Mathlib performance of \lh on \miniCTX-v2-test \citep{hu2025minictx} using the \modelsizethree model trained on Mathlib. For other settings than \settingfull, see \Cref{tab:minictx-performance-extended}.} \label{tab:minictx-performance}

\end{table}

% textbf has lower vertical space than paragraph
\textbf{\lh proves a significant number of theorems.} As shown in \Cref{tab:performance}, we find that \lh proves a significant proportion of test theorems, with 33.3\% proved by the \modelsizethree model in the \settingcumul setting, and 37.3\% when accumulated over model sizes. We also test giving ground-truth premises (those that appear in the human-written proof) to \lh, which serves as a theoretical best-case scenario of how \lh would perform if the models achieved 100\% recall, and this proves 43.0\% of the theorems. Compared to previous work, \lh approaches this limit in the settings considered.

\textbf{Performance scales with model size and accumulation.}
In \Cref{tab:performance} and \Cref{fig:performance}, as we increase our model size, for most settings the recall and proof rates also correspondingly increase (e.g., recall@32 increases from 67.8\% to 72.7\% and \settingfull proof rate increases from 27.9\% to 30.1\%). We also observe that by accumulating across different model sizes or taking the union of neural (our model) and symbolic (MePo) approaches, the proof rate increases much more than scaling the model alone (e.g., \settingfull proof rate increases to 34.5\% when accumulated), meaning different selectors prove different sets of theorems. More effective methods of ensembling models may be explored in future work.

\textbf{\lh settings offer different abilities at different costs.}
For the settings \settingauto, \settingaesop, \settingaesopauto, and \settingfull, the proof rate roughly increases in this order for all models. This shows that each part of the full pipeline incrementally contributes to the final proof rate. Their mean run times on Mathlib-test are {%
\sisetup{round-mode=figures, round-precision=2, number-unit-product=}%
\SI{4.3}{\second},
\SI{0.92}{\second},
\SI{4.9}{\second}, and
\SI{6.6}{\second}%
} respectively, so the non-\settingfull variants are computationally appealing alternatives that recover some of the \settingfull performance.
We also note that \settingcumul achieves higher proof rates than \settingfull, so some cases benefit from a partial pipeline (e.g.\ if the \settingfull pipeline does not terminate).

\textbf{\lh shows robust out-of-Mathlib generalization.} As shown in \Cref{tab:minictx-performance}, the performance on \miniCTX-v2-test \citep{hu2025minictx} is comparable to the performance on Mathlib---the proportion of theorems proven by \lh with the \modelsizethree selector, out of theorems proven with the ground-truth premises (i.e.\ the best-case scenario), is 73.5\% on Mathlib and 79.4\% on \miniCTX with the \settingfull pipeline, showing that performance does not decrease (the other settings also have comparable numbers; see \Cref{tab:minictx-performance-extended}). We also confirm in the table that if no premises are supplied, the performance is much worse (except for the Foundation split), which indicates that the \lh is not just proving trivial theorems.

Across all benchmarks, there are a handful of common patterns characterizing problems that \lh fails to solve. Some problems are not solved because \lp fails to retrieve necessary lemmas, as can be seen from the fact that the ground truth outperforms all other premise selectors in \Cref{tab:performance,tab:minictx-performance}. Some problems are not solved because they are out of scope for Lean-auto's translation procedure, which can occur when the problem in question contains features from dependent type theory not easily translated to higher-order logic. And some problems are not solved because the solutions require forms of reasoning not supported by Aesop, Zipperposition, or Duper (e.g.~induction or arithmetic). Comparatively, it is rare for \lh to succeed at proof search with Zipperposition but fail at the proof reconstruction stage with Duper. See \Cref{sec:proof-difficulty,sec:error-analysis} for additional analysis.

\subsection{Comparisons}

We compare \lp against the following existing work: non-LM methods MePo \citep{meng2009lightweight_mepo} and \citet{piotrowski2023machine}, and LM-based ReProver \citep{yang2023leandojo}. We use a recent adaptation of MePo from Isabelle to Lean (implemented by Kim Morrison), tune its parameters $p$ and $c$ on our evaluation recall@$k$, and apply our premise blacklist.
For \citet{piotrowski2023machine}, we select their random forest model with highest reported performance; in order to overcome errors, we modified its training and evaluation in a way that only gives them unfair advantage, so the reported performance is an upper bound. (See \Cref{sec:baseline-settings} for details of both methods.)
We find that \lp clearly outperforms either method---for the \modelsizethree model, our recall@32 is 73\% higher relative to MePo and our \settingcumul proof rate is 21\% higher (\Cref{tab:performance}). Meanwhile, the union of theorems our models and MePo can solve is much higher than each method separately, indicating that symbolic and neural methods have complementary strengths. We believe effective combinations of neural and symbolic methods warrant future investigation.

We retrain ReProver using their training and retrieval scripts, but on our train/valid/test splits and an updated Mathlib version (\Cref{sec:baseline-settings}).
\lp clearly outperforms ReProver \citep{yang2023leandojo} in terms of recall and proof rate---\lh using our \modelsizethree model (82M parameters) proves 150\% more theorems relative to using ReProver (218M) in the \settingfull setting and 50\% more in the \settingcumul setting. We attribute the performance gap to two main factors. First, ReProver focuses on premises used in the next tactic for tactic generation, while \lp focuses on finishing the entire proof, so the definitions of ground-truth premises are different (\Cref{sec:state-premise-extraction}). Second, \lh uses techniques such as term-style proof extraction, extraction of implicit premises, and better premise signature formatting (\Cref{sec:data-extraction}).
ReProver also uses an $\ell^2$ loss on the cosine similarity for training, rather than our contrastive loss, and we suspect this also contributes to our better performance.

\subsection{Ablations} \label{sec:ablation}

\begin{table}[!tbp]
\small

% control whether display valid or test performance
% display valid for ablations
\newcommand{\validortest}[2]{#1}

\centering
\adjustbox{max width=\linewidth}
{
% rounding and aligning numbers
\sisetup{
  round-mode=places, round-precision=1,  % 1 after .
  table-number-alignment=center,
  detect-weight=true, % detect-inline-weight=math % for bold
}
\begin{tabular}{
    l
    S[table-format=2.1]
    S[table-format=2.1]
    S[table-format=2.1]
    S[table-format=2.1]
    S[table-format=2.1]
    S[table-format=2.1]
    S[table-format=2.1]
}
\toprule
\textbf{Premise selector} & \multicolumn{2}{c}{\textbf{Recall} (\%)} & \multicolumn{5}{c}{\textbf{Proof rate} (\%)} \\
\cmidrule(lr){2-3} \cmidrule(l){4-8}
& {@16} & {@32} & {\settingaesop\unskip} & {\settingauto\unskip} & {\settingaesopauto\unskip} & {\settingfull\unskip} & {\settingcumul\unskip} \\
\midrule
\lp (\modelsizetwo)
& \validortest{61.1}{58.630562068158476}
& \validortest{71.9}{68.1119083849065}
& \validortest{29.8}{23.0924}
& \validortest{22.6}{20.0803}
& \validortest{30.2}{26.1044}
& \validortest{34.6}{28.5141}
& \validortest{37.6}{30.7229}
\\
\quad $+$ naive data
& \validortest{57.5}{54.01321693205108}
& \validortest{66.8}{64.90807523231048}
& \validortest{29.3}{25.3521}
& \validortest{20.0}{18.7123}
& \validortest{28.5}{25.5533}
& \validortest{33.1}{29.5775}
& \validortest{35.2}{32.9980}
\\
\quad $-$ negatives
& \validortest{51.8}{50.02100003141451}
& \validortest{59.5}{57.81935501365684}
& \validortest{28.6}{23.0924}
& \validortest{20.0}{15.8635}
& \validortest{28.4}{23.6948}
& \validortest{33.0}{25.9036}
& \validortest{36.8}{29.3173}
\\
\quad $-$ loss mask
& \validortest{59.1}{57.17922369304475}
& \validortest{69.6}{65.47900054842151}
& \validortest{29.4}{}
& \validortest{21.2}{}
& \validortest{29.0}{}
& \validortest{34.4}{}
& \validortest{38.4}{}
\\
\midrule
Ground truth
&
&
& \validortest{30.8}{27.7108}
& \validortest{32.0}{33.1325}
& \validortest{38.4}{37.7510}
& \validortest{41.2}{40.9639}
& \validortest{43.6}{42.9719}
\\
\quad $+$ naive data
&
&
& \validortest{31.2}{28.3702}
& \validortest{30.0}{31.7907}
& \validortest{37.0}{36.4185}
& \validortest{39.8}{40.8451}
& \validortest{42.4}{42.1687}
\\
% \multicolumn{2}{l}{\quad + naive data + blacklist}
% & 
% & \validortest{30.8000}{}
% & \validortest{30.4000}{}
% &
% & \validortest{39.6000}{}
% & \validortest{TODO}{}
% \\
\bottomrule
\end{tabular}
}
\caption{Ablation study of \lh with different training settings on Mathlib-\validortest{valid}{test}.} \label{tab:ablations}

\end{table}

\Cref{tab:ablations} shows the performance of \lh on Mathlib-valid with some components removed: (1) we use a naive data extraction script that (a) uses default pretty-printing options, (b) disables our premise blacklist, and (c) disables collection of premises from \verb|simp| or \verb|rw| calls; (2) we do not sample negative premises during training ($B^-=0$); and (3) we disable masking positive in-batch premises in the contrastive loss, i.e.\ the denominator of \Cref{eq:loss} being simply the sum over all $B(1+B^-)$ premises in batch. We observe these changes clearly degrade performance.
Specifically, our data extraction (\Cref{sec:data-extraction}) is specifically designed with Lean-auto translation in mind, and we observe that settings with Lean-auto have a lower performance with a naive data extraction script. We observe that randomly sampling negative premises and the loss mask (\Cref{sec:model-training}) improve performance.
(Although the \settingcumul proof rate of \lh without the loss mask is higher, we strongly believe this is due to random noise, because all individual settings give lower performances, the recall is lower, and proof rate has higher variance than recall.)

\section{Conclusion}
We developed \lp, a novel premise selection tool for a hammer in dependent type theory, and combined neural premise selection with symbolic automation to build \lh, the first domain-general hammer in Lean. With comprehensive experiments, we show that \lh is performant on Mathlib compared to baselines, and generalizes well to \miniCTX-v2. \lp and \lh are designed with accessibility for Lean users in mind, and lay down groundwork for future hammer-based neural theorem proving in Lean.

\section*{Acknowledgments}
We would like to thank Kim Morrison for early discussions about premise selection and implementing a premise selection API and the MePo selector in Lean core,
Jasmin Blanchette for proofreading the paper and giving suggestions,
Yicheng Qian for discussions about premise selection and building and improving Lean-auto,
and
Jannis Limperg, Mac Malone, and Joachim Breitner for answering our technical questions. Work partially supported by NSF Grant DMS-2434614 and a gift from Convergent Research.

\section*{Reproducibility statement}  % Does not count toward page limit
We make all code for data extraction, model training, evaluation, and API integration publicly available with an open-source license. Data extraction is open source at \url{https://github.com/cmu-l3/ntp-toolkit/tree/hammer}, training script is at \url{https://github.com/hanwenzhu/LeanHammer-training}, premise selection server is at \url{https://github.com/hanwenzhu/lean-premise-server}, and the Lean premise selection API is partly in Lean core and partly at \url{https://github.com/hanwenzhu/premise-selection}. We open source both \lp and \lh as tactics in Lean at \url{https://github.com/JOSHCLUNE/LeanHammer}. We also release the extracted data and all trained models and baselines.

\bibliography{iclr2026_conference}
\bibliographystyle{iclr2026_conference}

\appendix
\section{Data extraction example}
\label{sec:data-example}

We provide an example of proof state and premise extraction to illustrate the features of our data extraction pipeline. 

Consider the following theorem, \verb|AlgebraicGeometry.Scheme.RationalMap.mem_domain|, proven in the module \verb|Mathlib.AlgebraicGeometry.RationalMap|:
\begin{lstlisting}[language=lean]
lemma RationalMap.mem_domain {f : X ⤏ Y} {x} :
    x ∈ f.domain ↔ ∃ g : X.PartialMap Y, x ∈ g.domain ∧ g.toRationalMap = f :=
  TopologicalSpace.Opens.mem_sSup.trans (by simp [@and_comm (x ∈ _)])
\end{lstlisting}
We extract its proof data, including its intermediate proof states and premises used (\Cref{sec:state-premise-extraction}).
This is a term-style proof (the proof does not start with \verb|by|), and our data extraction extracts premises corresponding to the whole proof of the theorem, with state (note the pretty-printing options that disable notations and print full names):
\begin{lstlisting}[language=lean]
X Y : AlgebraicGeometry.Scheme
f : X.RationalMap Y
⊢ ∀ {x : ↑↑X.toPresheafedSpace},
    Iff (Membership.mem f.domain x) (Exists fun g => And (Membership.mem g.domain x) (Eq g.toRationalMap f))
\end{lstlisting}
and the ground-truth set of premises used, which is the union of premises appearing in the compiled proof term (e.g.\ \verb|exists_exists_and_eq_and| implicitly invoked by \verb|simp|) and appearing explicitly (e.g.\ \verb|and_comm| appearing in the argument of the \verb|simp| call): [\verb|Iff.trans|, \verb|exists_prop|, \verb|Eq.trans|, \verb|of_eq_true|, \verb|iff_self|, \verb|congr|, \verb|funext|, \verb|exists_exists_and_eq_and|, \verb|TopologicalSpace.Opens.mem_sSup|, \verb|propext|, \verb|and_comm|, \verb|congrArg|].

This set of premises are then filtered by a blacklist of common logical premises and other ineligible premises, which removes trivial theorems such as \texttt{iff\_self (p : Prop) : (p $\leftrightarrow$ p) = true}. This helps the selector to only focus on the smaller subset of premises that are meaningful for a hammer. The resulting list of premises is [\verb|exists_prop|, \verb|funext|, \verb|exists_exists_and_eq_and|, \verb|TopologicalSpace.Opens.mem_sSup|].

For tactic-style (sub)proofs, the state before each tactic is also collected. In the proof above, this is the state before the \verb|simp| call:
\begin{lstlisting}[language=lean]
X Y : AlgebraicGeometry.Scheme
f : X.RationalMap Y
x : ↑↑X.toPresheafedSpace
⊢ Iff
    (Exists fun u =>
      And (Membership.mem (setOf fun x => Exists fun g => Exists fun x_1 => Eq g.domain x) u) (Membership.mem u x))
    (Exists fun g => And (Membership.mem g.domain x) (Eq g.toRationalMap f))
\end{lstlisting}
This state has the same ground-truth premises as the state above, since they correspond to the same theorem proof.
Compare this data extraction with prior work \citep{yang2023leandojo} that only extract premises that explicitly occur in the next tactic in tactic-style proofs (which is sensible for their purpose of textual next-tactic generation but not for a hammer that needs all relevant premises to close the current goal).

Aside from extracting data from its proof, this theorem may be used as a premise for down-stream theorems. Therefore we also collect it as a premise (\Cref{sec:signature-extraction}), with pretty-printed signature as follows:\footnote{
We tried excluding the theorem name (here \texttt{AlgebraicGeometry.Scheme.RationalMap.mem\_domain}) in a premise signature, but preliminary ablation experiments did not show performance gains.
}
\begin{lstlisting}[language=lean]
theorem AlgebraicGeometry.Scheme.RationalMap.mem_domain {X Y : AlgebraicGeometry.Scheme} {f : X.RationalMap Y} {x : ↑↑X.toPresheafedSpace} : Iff (Membership.mem f.domain x) (Exists fun g => And (Membership.mem g.domain x) (Eq g.toRationalMap f))
\end{lstlisting}
Note the difference between this signature and the raw source code string at the start: pretty-printing notation shorthands like $\dashrightarrow$ and $\exists$ is disabled, names are expanded to full names, implicit types are added (such as the type of \verb|x|), previously declared variables like \verb|X| and \verb|Y| are included, and the proof is not included. This is because our signature printing is a function of the type of the premise and not its source string as in \cite{yang2023leandojo}. This gives the entire information of a premise while standardizing its signature printing.

\section{LeanHammer example}
\label{sec:leanhammer-example}

We provide an example of a proof produced by \lh to illustrate the features of our hammer pipeline.

Consider the following theorem, \verb|associated_gcd_right_iff|, proven in the module \verb|Mathlib.Algebra.GCDMonoid.Basic|:

\begin{minipage}{\textwidth}
\begin{lstlisting}[language=lean]
theorem associated_gcd_right_iff [GCDMonoid A] {x y : A} :
    Associated y (gcd x y) ↔ y ∣ x :=
  ⟨fun hx => hx.dvd.trans (gcd_dvd_left x y),
    fun hxy => associated_of_dvd_dvd (dvd_gcd hxy dvd_rfl) (gcd_dvd_right x y)⟩
\end{lstlisting}
\end{minipage}

The initial goal state produced by this theorem's signature is:

\noindent\begin{minipage}{.45\textwidth}
\begin{lstlisting}[language=lean, frame=tlrb, title=Initial goal state (as printed by VS Code)]
A : Type
inst¹ : CancelCommMonoidWithZero A
inst : GCDMonoid A
x y : A
⊢ Associated y (gcd x y) ↔ y ∣ x
\end{lstlisting}
\end{minipage}\hfill
\begin{minipage}{.5\textwidth}
\begin{lstlisting}[language=lean, frame=tlrb, title=String sent to premise selection server]
A : Type
inst¹ : CancelCommMonoidWithZero A
inst : GCDMonoid A
x y : A
⊢ Iff (Associated y (GCDMonoid.gcd x y)) (Dvd.dvd y x)
\end{lstlisting}
\end{minipage}

Note that there are slight differences between this goal state as printed by VS Code and as printed by our state extraction procedure. These differences serve to disambiguate constants and remove potentially overloaded notation, and are described in \Cref{sec:signature-extraction}.

Given this initial goal state, our premise selection server looks up premises accessible by this proof, which are premises either imported by the current module or defined earlier in the current module. Among these, it returns the following ordered list of 32 premises: 

\begin{multicols}{2}
\begin{enumerate}
% using \verb so that the prime ' is not pretty-printed by Latex as apostrophe ’
\item \verb|GCDMonoid.gcd_dvd_left|
\item \verb|dvd_gcd_iff|
\item \verb|GCDMonoid.dvd_gcd|
\item \verb|GCDMonoid.gcd_dvd_right|
\item \verb|Associated.dvd_iff_dvd_right|
\item \verb|Associated.dvd_iff_dvd_left|
\item \verb|gcd_comm'|
\item \verb|gcd_eq_zero_iff|
\item \verb|Associated.symm|
\item \verb|Associated.dvd|
\item \verb|gcd_zero_right'|
\item \verb|Associated.trans|
\item \verb|gcd_dvd_gcd|
\item \verb|Associated.refl|
\item \verb|associated_one_iff_isUnit|
\item \verb|gcd_zero_left'|
\item \verb|instDecompositionMonoidOfNonemptyGCDMonoid|
\item \verb|associated_of_dvd_dvd|
\item \verb|instNonemptyGCDMonoid|
\item \verb|associated_gcd_left_iff|
\item \verb|gcd_mul_lcm|
\item \verb|gcd_assoc'|
\item \verb|dvd_dvd_iff_associated|
\item \verb|gcd_mul_right'|
\item \verb|gcd_mul_left'|
\item \verb|GCDMonoid.gcd_mul_lcm|
\item \verb|dvd_gcd_mul_of_dvd_mul|
\item \verb|gcd_mul_dvd_mul_gcd|
\item \verb|Associated.mul_left|
\item \verb|gcd_one_right'|
\item \verb|mul_dvd_mul_iff_left|
\item \verb|gcd_pow_left_dvd_pow_gcd|
\end{enumerate}
\end{multicols}

As described in \Cref{sec:variations}, \lh uses $k_1 = 16$ premises supplied to Lean-auto, and $k_2 = 32$ for premise application rules, so the first 16 premises are sent to Lean-auto and all 32 of the above premises are added to Aesop as premise application rules. The proof that \lh discovers is equivalent to the following:

\begin{minipage}{\textwidth}
\begin{lstlisting}[language=lean]
theorem associated_gcd_right_iff [GCDMonoid A] {x y : A} :
    Associated y (gcd x y) ↔ y ∣ x := by
  apply Iff.intro -- Applied by Aesop
  · intro a -- Applied by Aesop
    duper [a, GCDMonoid.gcd_dvd_left, Associated.dvd_iff_dvd_left]
  · intro a -- Applied by Aesop
    apply associated_of_dvd_dvd -- Premise application (18)
    · duper [a, GCDMonoid.dvd_gcd, Associated.dvd, Associated.refl]
    · apply GCDMonoid.gcd_dvd_right -- Premise application (4)
\end{lstlisting}
\end{minipage}

In this proof, Aesop begins by transforming the initial goal into subgoals with the constructor introduction rule \verb|Iff.intro|. The first subgoal is provable using just the first 16 premises supplied by the premise selector, so after Lean-auto translates it into higher-order logic, Zipperposition reports that \verb|GCDMonoid.gcd_dvd_left| (premise 1) and \verb|Associated.dvd_iff_dvd_left| (premise 6) entail the first subgoal on their own. Then, since it is known that only these two premises are required to solve the first subgoal, these two premises can be passed to Duper on their own, and Duper is able to produce a proof for the first subgoal.

The second subgoal cannot be proven by Lean-auto and Zipperposition using just the first 16 premises, but Aesop sees that \verb|associated_of_dvd_dvd| (premise 18) can be applied directly. After it does so, two more subgoals are created, the first of which can once again be proven with Lean-auto, Zipperposition, and Duper (using a different subset of the first 16 premises), and the second of which can be proven with a direct application of \verb|GCDMonoid.gcd_dvd_right| (premise 4). Since \verb|GCDMonoid.gcd_dvd_right| is part of the first 16 premises, it would also be possible for this final subgoal to be proven using Lean-auto, Zipperposition, and Duper, but because direct premise applications are assigned a higher priority than invocations of Lean-auto (20\% as compared to 10\%, see \Cref{sec:variations}), Aesop discovers the simpler proof first.
\section{Baseline settings} \label{sec:baseline-settings}

The following baselines are used in our analysis. The specific setup details of each baseline are listed below:
\begin{itemize}[leftmargin=*]
\item None: No premises are supplied to \lh.
\item MePo \citep{meng2009lightweight_mepo}: A prototype implementation of MePo was recently adapted into Lean by Kim Morrison\hide{ at \url{https://github.com/leanprover/lean4/tree/mepo}}. We note that MePo is designed for selecting a much larger number of premises than what is typically optimal for \lh (\Cref{sec:finding-k}). We select the final $k$ premises selected by MePo, as we experimentally confirm that the final premises selected are the most relevant, while supplying too many premises to \lh would decrease its performance (\Cref{fig:finding-k}). We tuned the parameters $p$ and $c$ in the range $p\in\{0.5,0.6,0.7,0.8,0.9\}$ and $c\in\{0.9,1.2,2.4,3.6\}$ on our test data, and found that the setting leading to highest recall@$k$ is $p=0.6$ and $c=0.9$ (which effectively only runs the inner loop of MePo once). We also filter out ineligible premises and Lean language-related (e.g.\ metaprogramming) premises, similar to what we do in \Cref{sec:signature-extraction}. 
% \item $k$-NN \citep{piotrowski2023machine}: this model selects the most similar premises to the proof state by comparing their \emph{features}, which are the collection of symbols appearing in the type. We select their model with the highest reported performance, which is the model trained on data extracted with \verb|n+b| features. We also modify their training to train on all Mathlib data (including our evaluation data), because it is nontrivial to modify their code to follow our data splits. This means that the training data given to their model includes our test/valid theorems. This gives their model an \emph{unfair advantage} (especially for $k$-NN---the retrieved premises would simply be the ground truth for $k=1$ nearest neighbors), so our observations are an upper bound on their model performance. We encountered non-terminating data extraction on a small number of modules, so we set a limit of 1,000 seconds for data extraction on each module. We also increased the wall-clock limit during evaluation from 300 to 1,000 seconds, since their premise selection code take much longer than our neural selector. This still results in about 300 time-outs out of 500 test theorems, so we only report the average over the successful $\sim$200 theorems in \Cref{tab:performance}. (We also note these time-outs give a selection bias toward theorems that are potentially easier and do not result in time-out.)
\item Random forest \citep{piotrowski2023machine}: this is a random forest model based on \emph{features}, which are the collection of symbols appearing in the goal.  We select their model with the highest reported performance, which is the model trained on data extracted with \verb|n+b| features. We also modify their training to train on all Mathlib data (including our evaluation data), because it is nontrivial to modify their code to follow our data splits. This means that the training data given to their model includes our test/valid theorems. This gives their model an \emph{unfair advantage}, so our observations are an upper bound on their model performance. We encountered non-terminating data extraction on a small number of modules, so we set a limit of 1,000 seconds for data extraction on each module. We also encountered time-out and out-of-memory issues during premise retrieval (some using >30GB RAM for a single retrieval), so we also increased \lh time-out from 300 to 1,000 seconds. We did an additional pass to ensure we evaluate on as many theorems as possible. This still results in 300 premise retrieval errors out of 500 test theorems, so we only report the average over the successful $\sim$200 theorems in \Cref{tab:performance}.

We do not test on their $k$-NN model because it is reported to be worse than the random forest model, and including our evaluation theorems in the training data gives $k$-NN significant unfair advantage.
\item ReProver \citep{yang2023leandojo}: We run LeanDojo's data extraction script on our train/valid/test splits and Lean 4 version (version v4.16.0), and retrain their model on this data. We retrieve premises from accessible premises using their script. We note that our definition of ground-truth premises (premises used in the entire proof) is different from their definition (premises used in the next tactic), because we focus on finishing the goal and they focus on tactic generation, so there is discrepancy between recall@$k$ in \Cref{tab:performance} and \citet{yang2023leandojo}. (We manually verified that the recall@10 value using their definition is similar to what they report, at about $38\%$, so our re-training worked properly.)
\end{itemize}

\section{Additional results}

\subsection{Proof rate and run time}

\begin{figure}[htb]
    \centering
    \begin{subfigure}[t]{0.53\textwidth}
        \centering
        \includegraphics[width=\linewidth]{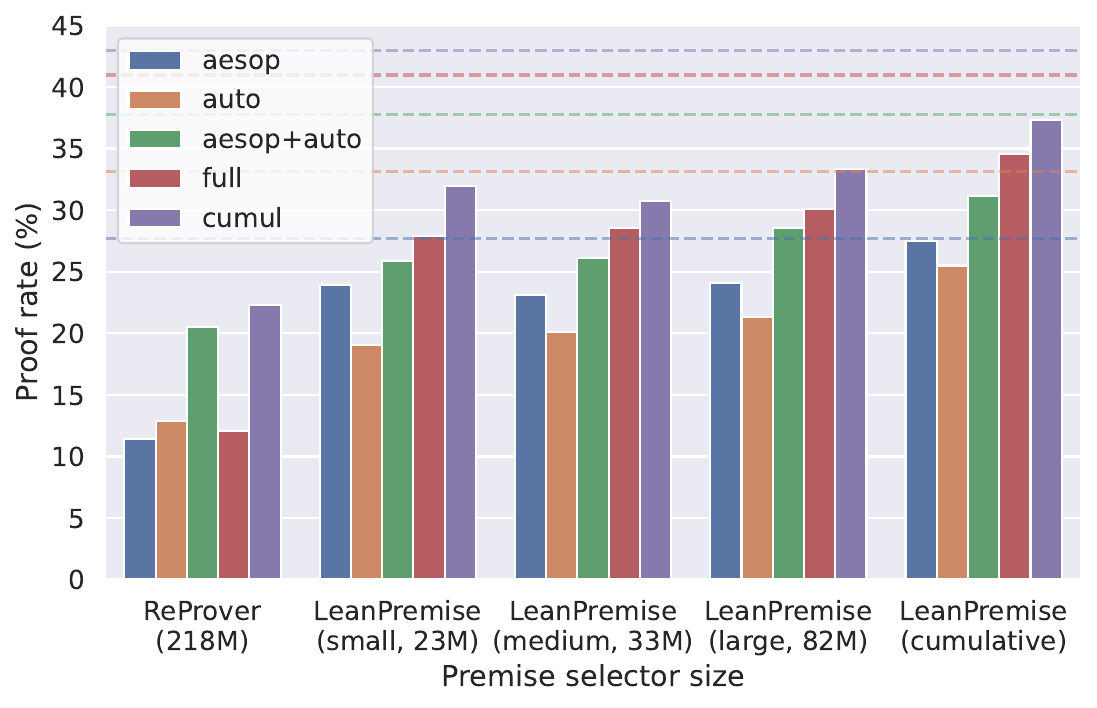}
        \caption{Proof rate on Mathlib-test using different premise selectors. Dashed lines are ground-truth proof rates.}
        \label{fig:performance}
    \end{subfigure}
    \hfill
    \begin{subfigure}[t]{0.45\textwidth}
        \centering
        \includegraphics[width=\linewidth]{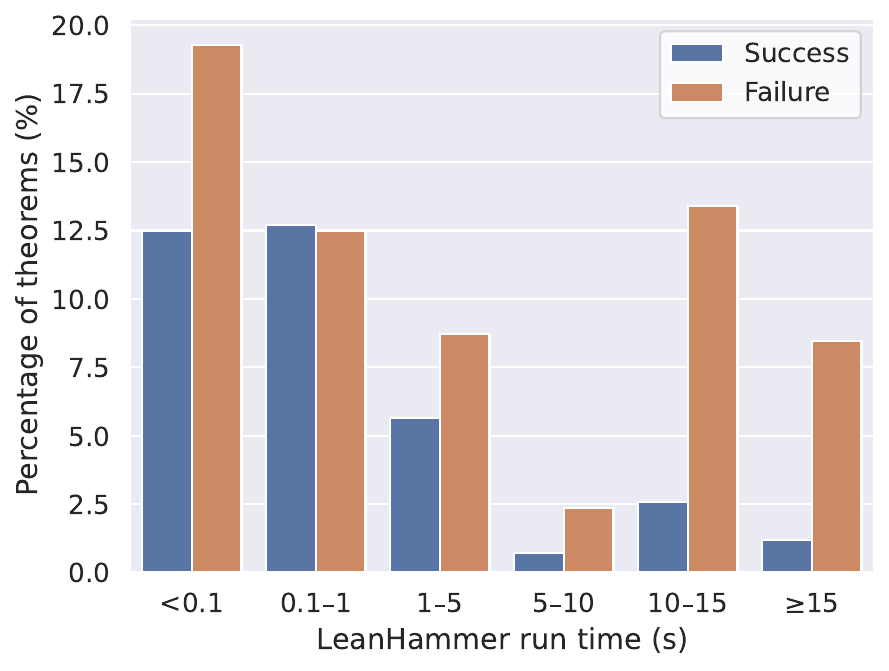}
        \caption{\lh (\settingfull) run time on Mathlib-test, depending on if the theorem was proven.}
        \label{fig:run-time}
    \end{subfigure}%
    \caption{Analysis of proof rate and execution speed in different settings.}
\end{figure}

In \Cref{fig:performance}, we show the performance of our models as the size of the premise selector scales on Mathlib-test. We note that using our models, performance scales well with model size.

In \Cref{fig:run-time}, we show the run time of \lh on Mathlib-test using the \modelsizethree model, depending on if the theorem was proven. Most successful applications of \lh run in under 1 second, but some theorems require a much longer time.

% Thomas: graphics are generated by the notebook https://colab.research.google.com/drive/1uljc8k8Z0FBHzIkGXw1yV7CuPn-qx_Iq#scrollTo=dsO2ZiA7qc7U
% (feel free to edit the graphics)

\subsection{Extended \miniCTX-v2 results}
\begin{table}[!tbp]
\centering
\adjustbox{max width=\linewidth}{
\sisetup{
  round-mode=places, round-precision=1,  % 1 after .
  table-number-alignment=center,
  detect-weight=true, % detect-inline-weight=math % for bold
}
\begin{tabular}{
    lc
    S[table-format=2.1]
    S[table-format=2.1]
    S[table-format=2.1]
    S[table-format=2.1]
    S[table-format=2.1]
    S[table-format=2.1]
    S[table-format=2.1]
}
\toprule
\textbf{Premise selector} & \textbf{Setting} &
\multicolumn{7}{c}{\textbf{Proof rate} (\%)} \\
\cmidrule(l){3-9}
&& {Carleson} & {ConNF} & {FLT} & {Foundation} & {HepLean} & {Seymour} & {Average} \\
\midrule
None & \settingaesop & 0.0 & 10.0 & 27.3 & 38.0 & 8.0 & 6.0 & 14.9
\\
\lp (\modelsizethree) & \settingaesop & 0.0 & 16.0 & 39.4 & 38.0 & 10.0 & 20.0 & 20.6
\\
Ground truth & \settingaesop & 2.4 & 16.0 & 30.3 & 40.0 & 12.0 & 22.0 & 20.4
\\
\midrule
None & \settingauto& 0.0& 10.0& 12.1& 32.0& 4.0& 4.0& 10.4
\\
\lp (\modelsizethree) & \settingauto & 0.0 & 10.0& 15.2 & 32.0& 4.0& 10.0 & 11.9
\\
Ground truth & \settingauto & 4.8 & 10.0 & 24.2 & 34.0 & 12.0 & 32.0 & 19.5
\\
\midrule
None & \settingaesopauto& 0.0 & 10.0& 27.3& 38.0& 8.0& 6.0& 14.9
\\
\lp (\modelsizethree) & \settingaesopauto & 0.0 & 10.0 & 30.3 & 40.0 & 10.0 & 16.0 & 17.7
\\
Ground truth & \settingaesopauto & 4.8 & 10.0 & 36.4 & 40.0 & 18.0 & 30.0 & 23.2
\\
\midrule
None & \settingfull
& 0.0 & 10.0 & 27.3 & 38.0 & 8.0 & 6.0 & 14.9
\\
\lp (\modelsizethree) & \settingfull
& 0.0 & 16.0 & 36.4 & 38.0 & 10.0 & 24.0 & 20.7
\\
Ground truth & \settingfull
& 7.1 & 16.0 & 39.4 & 40.0 & 20.0 & 34.0 & 26.1
\\
\midrule
None & \settingcumul & 0.0 & 10.0& 27.3& 38.0& 8.0& 6.0& 14.9
\\
\lp (\modelsizethree) & \settingcumul & 0.0 & 16.0 & 39.4 & 40.0 & 12.0 & 26.0 & 22.2
\\
Ground truth & \settingcumul & 7.1 & 16.0 & 39.4 & 40.0 & 20.0 & 38.0 & 26.8
\\
\bottomrule
\end{tabular}
}
\caption{Extended table of performance of \lh on each split of \miniCTX-v2-test \citep{hu2025minictx} using the \modelsizethree model trained on Mathlib.} \label{tab:minictx-performance-extended}

\end{table}

We present the complete results over all settings on \miniCTX-v2 in \Cref{tab:minictx-performance-extended}. We note that over all settings, the ratio of our proof rate to the proof rate given ground-truth premises is largely preserved, or even increases for some settings, from Mathlib to \miniCTX-v2.
For the Carleson split, our pipeline resulted in an unexpected error which we believe is fixable, but in the meantime we put the proof rate as 0.0 (as a lower bound of performance).

\subsection{Finding optimal \texorpdfstring{$k$}{k}} \label{sec:finding-k}

\begin{figure}[!htbp]
\centering
\includegraphics[width=0.5\linewidth]{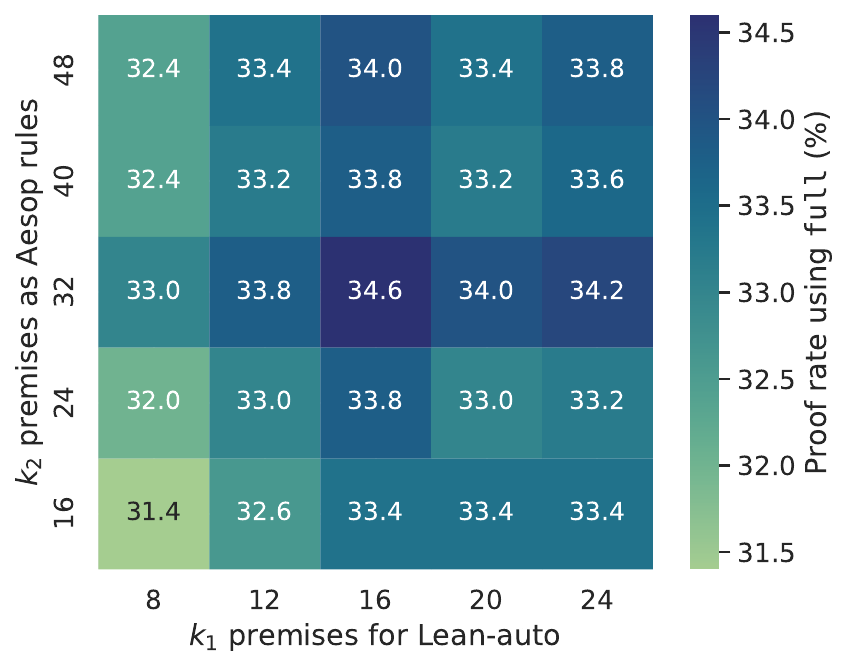}
\caption{Proof rate on Mathlib-valid by number of retrieved premises under \settingfull setting.}
\label{fig:finding-k}
\end{figure}

In order to determine the number $k$ of premises to retrieve and supply to \lh, we perform a sweep of possible numbers (\Cref{fig:finding-k}) and determine that $k_1=16$ premises should be supplied to Lean-auto and $k_2=32$ premises should be supplied for premise application rules. Their respective Aesop priority values 10\% and 20\% are determined similarly, though their effect is much less than changing $k$, so we omit the details.

\subsection{Analysis of proof rate by theorem difficulty}
\label{sec:proof-difficulty}

\begin{figure}[htb]
    \centering
    \begin{subfigure}[t]{0.9\linewidth}
        \centering
        \includegraphics[width=0.9\linewidth]{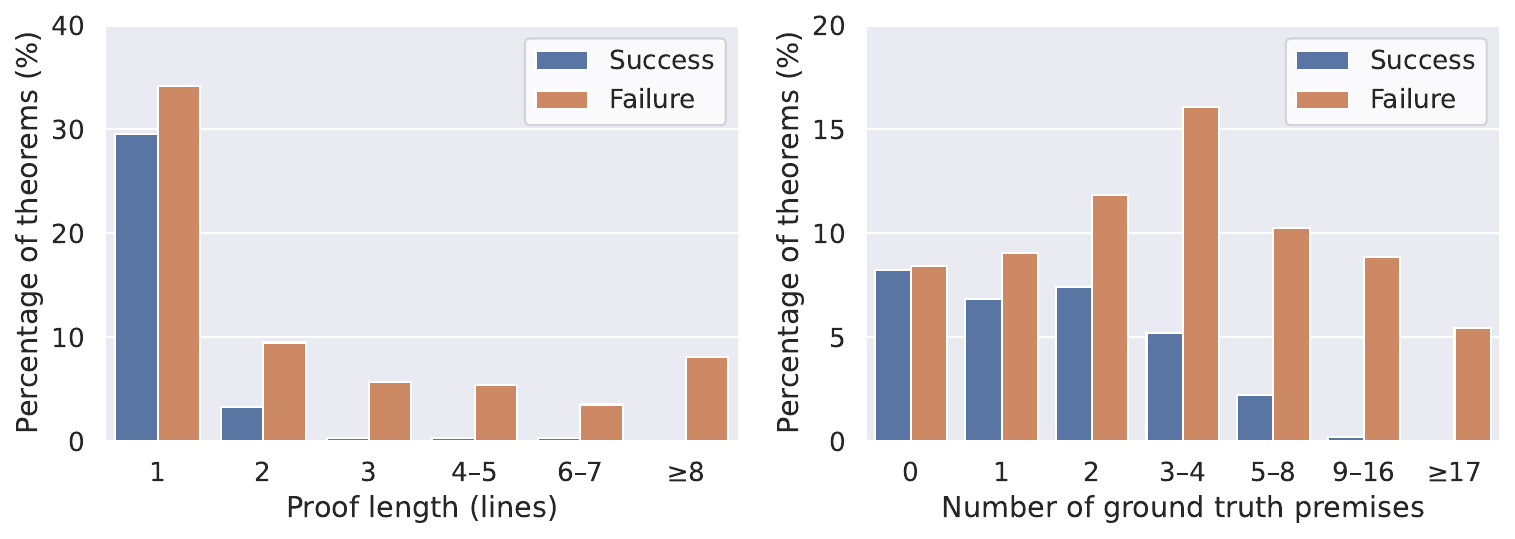}
        \caption{Theorem statistics by difficulty using the \modelsizethree model.} \label{fig:proof-difficulty-large}
    \end{subfigure}
    \begin{subfigure}[t]{0.9\linewidth}
        \centering
        \includegraphics[width=0.9\linewidth]{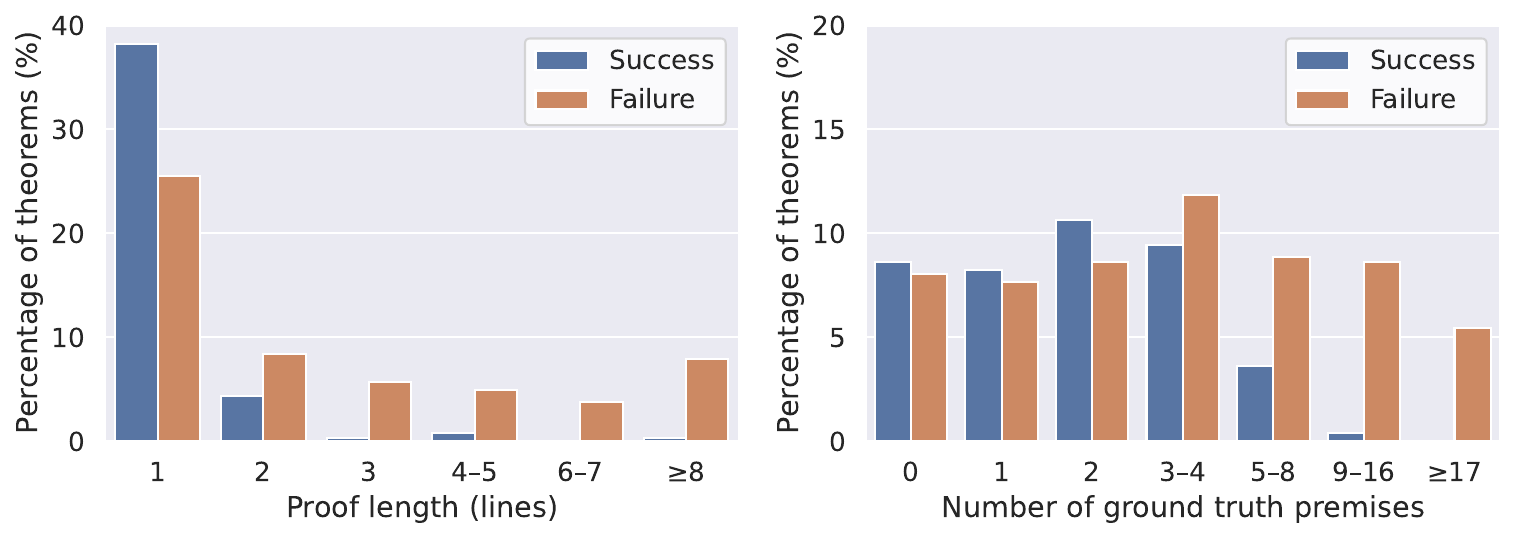}
        \caption{Theorem statistics by theorem difficulty using ground-truth premises.} \label{fig:proof-difficulty-gt}
    \end{subfigure}
    \caption{Analysis of theorem statistics by difficulty, on Mathlib-test with the \settingfull setting, depending on if the theorem was proven.} \label{fig:proof-difficulty}
\end{figure}

For each theorem in Mathlib-test, we record the number of lines of the human-written proof uses\footnote{Excluding proof headers such as \texttt{:= by} and \texttt{where}.} and the number of (filtered) premises used by the human-written proof, as proxy metrics of the difficulty of the theorem. Shown in \Cref{fig:proof-difficulty}, we found that almost all theorems that \lh solves, whether using our model or with ground-truth premises, use 1--2 lines in the human-written proof, while theorems not solved have a longer tail distribution. This means that, even if our premise selector were optimal, we would expect \lh to primarily be useful for solving the final few lines or small gaps in a proof. Note that proofs in Mathlib are often ``golfed'', and 1--2-line proofs still have a wide range of difficulties. Similarly, the number of ground-truth premises of theorems that \lh proves is usually no more than 8. These result imply that \lh is good at filling in the small gaps in proofs, as the search space becomes prohibitively large for longer proofs.

\subsection{Error analysis}
\label{sec:error-analysis}
The \lh pipeline has multiple components, and each part may encounter an error during a proof attempt. In order to identify parts that may be improved in the future, we record the source of error leading to each unsuccessful proof, specifically in the \settingauto setting (the part involving premise selection, Lean-auto translation, Zipperposition proof search, and Duper proof reconstruction), because the Aesop part is more established and well-understood \citep{limperg2023aesop}. When the ground truth premises (resp.\ premises retrieved by the \modelsizethree model) were supplied to \lh, the results are as follows on Mathlib-test:
\begin{enumerate}[leftmargin=*]
\sisetup{round-mode=places, round-precision=1, number-unit-product=}
\item \SI{21.68674699}{\percent} (\SI{26.70682731}{\percent}) of the theorems could not be translated by Lean-auto into the TH0 format used by Zipperposition (usually because the theorem itself or one of the premises supplied is outside the scope of the current translation procedure).
\item \SI{43.57429719}{\percent} (\SI{50.40160643}{\percent}) of the theorems were translated by Lean-auto, but could not be proven by Zipperposition. This may be because necessary premises were not retrieved, Zipperposition was unable to perform the required form of reasoning (e.g. arithmetic or induction), or the translation did not preserve enough information (e.g. because the translation did not unfold some necessary constant).
\item \SI{1.606425703}{\percent} (\SI{1.204819277}{\percent}) of the theorems were proven by Zipperposition, but its proof could not be reconstructed in Lean by Duper.
\item \SI{0}{\percent} (\SI{0.4016064257}{\percent}) of the theorems encountered another error.
\item The remaining \SI{33.1325}{\percent} (\SI{21.28514056}{\percent}) were successfully proven.
\end{enumerate}
This shows that there may be improvements gained from (1) increasing recall@$k$ of our premise selector, (2) improving translation of Lean into TH0, and (3) incorporating complementary tactics such as \texttt{grind} capable of solving problems not ideally suited to Duper, Zipperposition, or Aesop (e.g. problems involving arithmetic).

\end{document}